\documentclass[letterpaper, 10 pt, conference]{ieeeconf}  % Comment this line out if you need a4paper

\IEEEoverridecommandlockouts                              % This command is only needed if
\makeatletter
\let\NAT@parse\undefined
\makeatother
\usepackage[numbers,sort&compress]{natbib}

\usepackage[pdftex]{graphicx}
\usepackage{amsmath}
\usepackage{amssymb}  % assumes amsmath package installed
\usepackage{multirow}
\usepackage{array,booktabs}
\usepackage{diagbox}
\usepackage{balance}
\usepackage{xspace}
\usepackage{pifont}
\usepackage{algorithm}
\usepackage{algpseudocode}
\usepackage{adjustbox}
\usepackage{romannum}
\usepackage{caption}
\usepackage{subcaption}
\usepackage[final]{hyperref}
\hypersetup{
 colorlinks=true,
 linkcolor=blue,
 filecolor=magenta,
 urlcolor=blue,
 citecolor=black
}
\usepackage{cleveref}
\crefname{figure}{Fig.}{Figs.}
\Crefname{figure}{Fig.}{Figs.}
\usepackage[font=small]{caption}
\usepackage{rpm_SIunits}
\usepackage{rpm_acronyms}
\usepackage{rpm_math}
\usepackage{rpm_misc}

\usepackage{soul,color}
\usepackage{lipsum}

\usepackage{xcolor}

\usepackage{tikz} % \checkmark
\usepackage{subcaption}
\usepackage{caption}
\title{\LARGE \bf TIR-Diffusion: Diffusion-based Thermal Infrared Image Denoising\\ via Latent and Wavelet Domain Optimization}     

\author{Tai Hyoung Rhee${}^{1}$, Dong-Guw Lee${}^{1}$, and Ayoung Kim${}^{1*}$
\thanks{$^\dagger$This work was supported by the National Research Foundation of Korea(NRF) grant funded by the Korea government(MSIT) (No. RS-2023-00241758).}
\thanks{$^{1}$T. Rhee and A. Kim are with the Dept. of Mechanical Engineering, SNU, Seoul, S. Korea {\tt\small [williamrhee, donkeymouse, ayoungk]@snu.ac.kr}}%
}

\begin{document}

% \makeatletter
%   \let\@oldmaketitle\@maketitle% Store \@maketitle
%   \renewcommand{\@maketitle}{\@oldmaketitle% Update \@maketitle to insert...
%   % \bigskip
%   \centering
%   \vspace{+3mm}
%     \includegraphics[trim= 0cm 0cm 0cm 0cm, clip,width=0.8\textwidth]{figures/figure1.pdf}
%     \captionof{figure}{Outline of Denoising Method conducted on Multi-spectral Dataset \cite{dai2021multispectraldataset}}
%     }
    
%     \label{fig:main}

\makeatother

%\onecolumn
\maketitle
\thispagestyle{empty}
\pagestyle{empty}

% \begin{figure*}[h]
%     \centering
%     \includegraphics[width=\textwidth]{figures/figure1.pdf}
%     \caption{Denoising results conducted on Multi-spectral Dataset \cite{dai2021multispectraldataset}}
%     \label{fig:figure1}
% \end{figure*}

\begin{abstract}

Thermal infrared imaging exhibits considerable potentials for robotic perception tasks, especially in environments with poor visibility or challenging lighting conditions. However, TIR images typically suffer from heavy non-uniform fixed-pattern noise, complicating tasks such as object detection, localization, and mapping. To address this, we propose a diffusion-based TIR image denoising framework leveraging latent-space representations and wavelet-domain optimization. Utilizing a pretrained stable diffusion model, our method fine-tunes the model via a novel loss function combining latent-space and \ac{DWT} / \ac{DTCWT} losses. Additionally, we implement a cascaded refinement stage to enhance fine details, ensuring high-fidelity denoising results. Experiments on benchmark datasets demonstrate superior performance of our approach compared to state-of-the-art denoising methods. Furthermore, our method exhibits robust zero-shot generalization to diverse and challenging real-world TIR datasets, underscoring its effectiveness for practical robotic deployment.
%\ak{Overall well-written. The introduction and the very first sentence should be improved.}

\end{abstract}

\section{Introduction}
\label{sec:intro}

Thermal infrared (TIR) imaging has emerged as a transformative technology for robotic perception, unlocking unprecedented capabilities in challenging conditions where conventional vision sensors falter, such as darkness, smoke, or heavy occlusion. Applications such as autonomous surveillance, search-and-rescue, and inspection in hazardous environments particularly benefit from TIR’s ability to capture emissive temperature patterns. Despite these advantages, TIR data are often plagued by significant noise resulting from sensor design constraints and thermal fluctuations in real-world conditions~\cite{gil2024fieldscale}. Such noise not only degrades visual clarity but also impedes crucial downstream tasks in robotics, including object detection, collision avoidance, and simultaneous localization and mapping.

Recent advances in generative modeling have significantly expanded the methodological landscape for TIR image restoration. Classical approaches relied on spatial domain filters such as bilateral filtering or transform domain techniques including wavelet shrinkage optimized via bio-inspired algorithms \cite{liu2017classicalimagedenoising}, which often struggled to preserve fine textural details while suppressing TIR-specific non-uniform noise patterns. With the advent of deep learning techniques, widely implemented vision networks were utilized for the task of TIR image denoising, including convolutional neural networks \cite{he2018single_image_based_nonuniformity, li2021infrared, tang2024thermal}, and adversarial networks \cite{liu2025deal, guo2024lightweight}.

\begin{figure}[H]
    \centering
    \includegraphics[width=1.0\columnwidth]{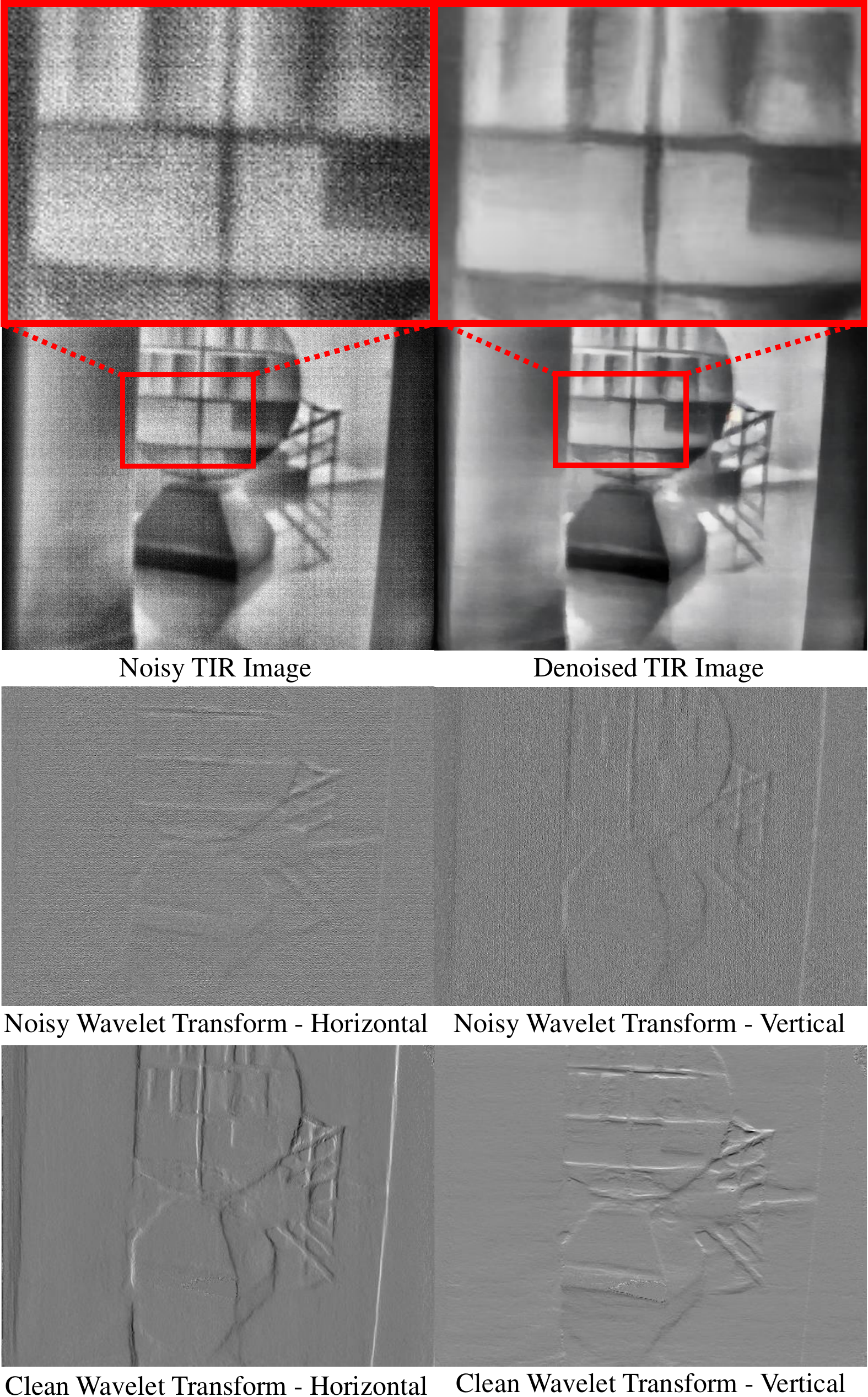}
    \caption{Outline of denoising method conducted on Multi-spectral Dataset \cite{dai2021multispectraldataset}. Wavelet transform robustly distinguishes and separates severe non-uniform fixed-pattern noise (FPN) specific to TIR images. Using input noisy-clean image pairs, discrete wavelet coefficients are computed, and their discrepancies are learned, enabling effective TIR image denoising.}
    \label{fig:main}
    \vspace{-5mm}
\end{figure}

However, existing works exhibit two main drawbacks: the absence of real noise data and the paucity of training data. Obtaining real clean-noisy TIR image pair is difficult due to the complex and laborious methodology it requires, leading to previous works employing synthetic Gaussian \cite{guo2024lightweight} or strip \cite{liu2025deal,liu2023thermal_tv-dip} noise added on to clean TIR images. Models utilizing synthetic Gaussian noise training data are incapable of denoising complex fixed-pattern noise (FPN), where as the discrepancy between synthetic and real strip noise hinders the models' ability to recognize real noise. With works utilizing real images \cite{liu2025twostage}, the amount of data is insufficient to train a deep neural network from the ground up, inducing suboptimal performance.

%The advent of deep learning introduced targeted denoising architectures like Deformable Denoising CNNs (DeDn-CNN) \cite{tang2024thermal}, which employed deformable convolutions to better adapt to irregular noise distributions in electrical equipment thermal inspections. Subsequent developments explored adversarial training frameworks, exemplified by lightweight GANs utilizing structural reparameterization and infrared-optimized transfer learning strategies to enhance feature representation in small-sample scenarios \cite{guo2024lightweight}. However, current methodologies frequently exhibit limitations in robotics applications where preserving micron-level thermal gradients is critical – either inheriting visible-spectrum bias from transfer learning architectures or relying on single-step neural filters that inadequately model the spatially variant noise characteristics intrinsic to uncooled microbolometer sensors.

In this work, we introduce TIR-Diffusion—a large foundational vision model–based TIR denoising framework that repurposes a pretrained diffusion U-Net to address the challenges posed by noisy thermal imagery. Real clean-noisy TIR image pairs are used for tine-tuning, while a large pretrained model is utilized to overcome the lack of training data. Diffusion models provide a highly diverse variety compared to other generative models, but requires abundant training data. The lack of data acted as a bottleneck preventing existing works from implementing the diffusion model, whereas we incorporated its full capacity via pretrained model implementation. Additionally, a novel approach of wavelet transform is used, utilizing its capability to distinguish between the image and noise for real TIR images as shown in \figref{fig:main}. For enhanced output image fidelity, we introduce an expansive cascaded model, leading to the best performance on testing data and zero-shot performance on unseen datasets compared to existing work.  

%Using the original Variational Autoencoder (VAE) from Stable Diffusion \cite{stablediffusion}, both the ground-truth and noisy TIR images are encoded into latent representations. A custom conditioning projection module transfroms noisy latent tensor into a cross-attention embedding, where its purpose is to guide the U-Net's denoising process. The U-Net takes the noisy latent tensor, conditioning tokens and a random timestep as input, which outputs a predicted denoised latent tensor compared with the ground truth via a combined loss incorporating latent representation and 2D discrete wavelet transforms. Additionally, for enhancing fine details, a pixel-wise model of identical architecture is cascaded, improving general performance. Together, these elements form a robust denoiser that preserves fine details in TIR images while mitigating heavy noise. Empirical results show improved performance compared to standard CNN-based methods, especially regarding high-frequency fidelity and edge sharpness.

The main contributions are summarized as follows:
\begin{itemize}
    \item \textbf{Large Pretrained Diffusion Model to Mitigate Data Scarcity and Enhance Diversity}: With no existing works implementing the diffusion model due to scarcity of data, we utilize the high performance and various advantages of a large foundational model by repurposing the Stable Diffusion model \cite{stablediffusion} for TIR denoising. This helps addressing the limited availability of high-quality TIR data while enhancing sample diversity provided by the diffusion model, leading to improved generalization and robustness in real-world scenarios.
    %\ak{Did anyone else use the diffusion for TIR? What is our novelty beyond them?}
    
    \item \textbf{Utilizing Wavelet Transform and Unique Models Designed Specifically for Thermal Image Denoising}: With existing works computing loss on pixel-level or latent space, we propose a combined loss function that unifies latent-space and wavelet-domain constraints to balance global structure preservation and fine-detail enhancement, with an additional cascaded pixel-level model for improved clarity and fidelity. From experiments, we observed the ability of wavelet transform to capture real noise of TIR images.
    %\ak{The same thing. What is the novelty we have by using the wavelet? You should be more explicit.}
        
    \item \textbf{Robust Zero-shot Performance}: We demonstrate scalable, high-quality denoising performance on diverse TIR datasets relevant to real robotic tasks, paving the way for more robust thermal perception in challenging environments. The model excels even in datasets completely different from its training set. 
    
\end{itemize}

\begin{figure*}
    \centering
    \includegraphics[width=1\linewidth]{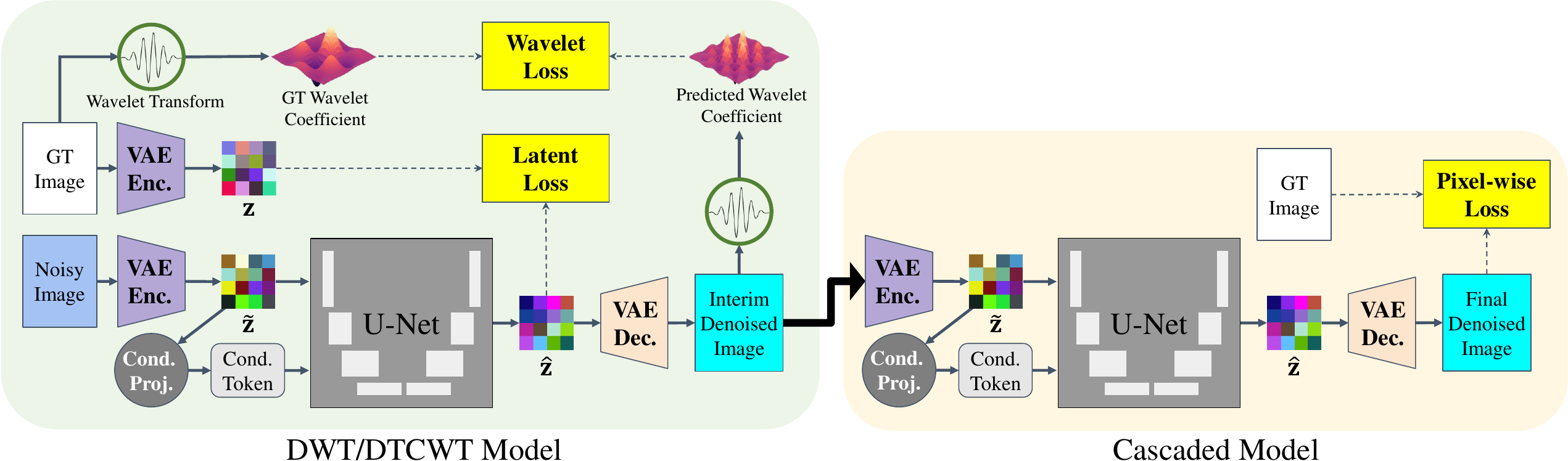}
        \vspace{+0.3mm}
    \caption{Diagram for DWT/DTCWT Model and Cascaded Model. The input noisy image is initially denoised by the DWT/DTCWT model, trained on the latent and wavelet domain. If cascading model is selected, the output interim denoised image from the DWT/DTCWT is inputted to the cascaded model, sharing the same Stable Diffusion \cite{stablediffusion} architecture as the DWT/DTCWT model, but trained on pixel-level.}
    \label{fig:model_diagram}
        \vspace{-3mm}
\end{figure*}

\section{related work}
\label{sec:relatedwork}

\subsection{Classical Image Denoising}

Image denoising is researched and utilized in various fields and domains, not only including TIR images but also RGB, medical images and more. Prior to the rise of deep learning techniques, traditional methods were prevalent, mainly categorized as spatial domain filtering and transform-based techniques.

Spatial domain approaches, such as Gaussian smoothing, median filtering, and bilateral filtering \cite{buades2005image_denoising_review} \cite{zhang2008multiresolution_bilateral_filtering}, are among the earliest and most widely adopted methods for image denoising. Gaussian filters reduce noise by averaging neighboring pixel intensities but tend to blur critical image details, which can hinder robotic perception systems that rely on image sharpness. Median filtering is robust against impulse noise and preserves edges more effectively; however, it may introduce artifacts or lose fine texture details in complex scenes. Bilateral filters, which consider both spatial proximity and intensity similarity, have shown promise in preserving edges while reducing noise, although their performance can deteriorate in highly textured or low-contrast TIR images typical in robotics scenarios.

Transform-based denoising techniques, particularly those employing wavelet transforms \cite{pathak2009the_wavelet_transform}, have been extensively studied in image processing. Wavelet-based methods exploit the multiscale decomposition capabilities of wavelets to separate noise components from meaningful features, enabling selective noise removal while preserving critical information. Discrete Wavelet Transform (DWT) methods \cite{skodras2003discrete_dwtintroduction} have been successful due to their computational efficiency and strong noise suppression capabilities. However, DWT approaches can sometimes produce artifacts near edges and singularities in the image, prompting researchers to investigate more sophisticated methods such as the Dual-tree Complex Wavelet Transform (DTCWT) \cite{dtcwt}. DTCWT offers improved directional selectivity and shift invariance, resulting in better preservation of edges and textures, which is highly beneficial for robotic vision applications requiring precise feature extraction and object localization. 

A more recent method proposed by \citet{liu2017classicalimagedenoising} combines the DTCWT with improved Fruit Fly Algorithm for noise variance threshold optimization and bilateral filter for re-processing, enhancing its performance. Such methods excelled at removing Gaussian noise, but still lacked in performance when denoising TIR images with large FPN and strip noise.

\subsection{Deep Learning-Based Thermal Image Denoising}

Advancement of deep learning has significantly advanced TIR image denoising, addressing limitations of traditional Gaussian methods.  \citet{he2018single_image_based_nonuniformity} and TV-DIP \cite{liu2023thermal_tv-dip} introduced a CNN model trained from simulated column FPN, but showed limitations due to reliance on artificially generated pairs. \citet{saragadam2021deepir} proposed DeepIR, a physics-inspired network requiring multiple images from camera jitter, limiting single-image applicability. \citet{guo2024lightweight} developed a lightweight GAN-based method using phased transfer learning, optimized mainly for additive white Gaussian noise rather than diverse thermal degradations. \citet{liu2025twostage} employed an encoder-decoder network similar to that of FFDNet \cite{zhang2018ffdnet} and frequency decomposition for single-image restoration, but their reliance on paired datasets and specific network designs restricted generalization. Concurrently, the DEAL framework \cite{liu2025deal} dynamically simulated TIR degradations via adversarial learning, yet its performance depended heavily on minimal training datasets.

Our method addresses these shortcomings by repurposing a large pretrained diffusion model, effectively overcoming data scarcity and enhancing diversity for better generalization. Furthermore, by integrating wavelet transform and designing models specifically for TIR denoising, our combined latent-space and wavelet-domain constraints more accurately capture real thermal noise, resulting in improved clarity, fidelity, and robustness in practical scenarios.

\section{Methods}

%---------------------------------------------------------------
\subsection{Preliminaries}

\subsubsection{Wavelet Transform}
Wavelet transform is a mathematical technique used for signal analysis at multiple scales, offering both time and frequency localization. It decomposes a signal into a set of wavelet functions, which are generated by scaling and shifting a mother wavelet. This allows for a multi-resolution analysis of the signal, capturing various frequency components at different scales. Mathematically, the wavelet transform of a signal $x(t)$ is given by

\begin{equation} Y(a,b) = \frac{1}{\sqrt{a}}\int_{-\infty}^{\infty} x(t)    \psi \left( \frac{t-b}{a} \right)  dt \end{equation}

where $\psi$ is the orthonormal wavelet function with scaling factor $a$ and time-shift factor $b$. \\

\subsubsection{2D Discrete Wavelet Transform}
For discrete data, namely images, the 2D discrete wavelet transform (DWT) applies a discrete set of scaling and shifting operations on both rows and columns to represent the image at different levels of resolution. For discrete cases, the wavelet function becomes

\begin{equation}
    \psi_{j,l}(t) = \frac{1}{\sqrt{a}} \psi \left( \frac{t-b}{a} \right) = 2^{-j/2} \psi \left( 2^{-j} t - l \right)
\end{equation}

\noindent where $\psi_{j,l}(t)$ is the scaled and shifted version of the mother wavelet $\psi(t)$. \cite{pathak2009the_wavelet_transform} The wavelet function can be interpreted as a bandpass filter \cite{skodras2003discrete_dwtintroduction} at a specific scale, which can be decomposed into a combination of a low-pass and high-pass filters.

Low-pass filters $h[\cdot]$ and high-pass filters $g[\cdot]$ are combinationally applied to the rows and the columns of the image in order, each followed by a subsampling of 2, where the results of 4 different combinations leads to 4 coefficients. Ordering each combination with L and H representing low-pass and high-pass filter, respectively, LL leads to low-pass components $Y^{lo}$, LH leads to horizontal components $Y^h$, HL leads to vertical components $Y^v$, and HH leads to diagonal components $Y^d$ of the wavelet transform \cite{pytorch_wavelets}. Each components at point $(n,m)$ on the image can be mathematically expressed as

\begin{subequations}
    \begin{equation}
        Y^{lo}_{j}[n,m] = \sum_k \sum_l f[k,m] \cdot h_x[n-k] \cdot h_y[1-m]
    \end{equation}
    \begin{equation}
        Y^h_{j}[n,m] = \sum_k \sum_l f[k,m] \cdot g_x[n-k] \cdot h_y[1-m]
    \end{equation}
    \begin{equation}
        Y^v_{j}[n,m] = \sum_k \sum_l f[k,m] \cdot h_x[n-k] \cdot g_y[1-m]
    \end{equation}
    \begin{equation}
        Y^d_{j}[n,m] = \sum_k \sum_l f[k,m] \cdot g_x[n-k] \cdot g_y[1-m]
    \end{equation}
\end{subequations}

where $j$ and $l$ are indices representing the level of decomposition and the transition index in the discrete domain. The specific equation of the filters depend on the type of wavelet, which will be explained in \secref{ssec:comparing_wavelets}. \\

\subsubsection{Dual-Tree Complex Wavelet Transform}

The DTCWT is an advanced extension of the traditional wavelet transform that improves upon several aspects of the DWT, particularly in areas including directionality, shift-invariance, and the handling of complex signals. DTCWT was introduced to achieve better performance in applications such as image processing, compression, and signal analysis, where both multiresolution and directional selectivity are required.

The key idea behind DTCWT is to use two separate wavelet trees for the decomposition, a real wavelet and an imaginary wavelet, which are designed to provide complex-valued coefficients and better angular resolution. The two wavelet transforms operate in parallel, producing complex coefficients at each decomposition level, allowing DTCWT to offer better directionality and phase information. The mathematical formulation is nearly identical to that of DWT, with the only difference of having a real part and an imaginary part. For instance, the mathematical expression of the horizontal component $Y^h$ is expressed as

\begin{align}
    Y^h_{j}[n,m] &= \underbrace{\sum_k \sum_l f[k,m] \cdot g_x^{(r)}[n-k] \cdot h_y^{(r)}[1-m]}_{\text{Real Part}} 
    \notag \\
    & +\space i \underbrace{\sum_k \sum_l f[k,m] \cdot g_x^{(i)}[n-k] \cdot h_y^{(i)}[1-m]}_{\text{Imaginary Part}}
\end{align}

where $(r)$ and $(i)$ superscript denotes the real and imaginary filters, which varies with the angle of analysis. Unlike the DWT allowing only horizontal, vertical and diagonal, DTCWT allows various angles of analysis, where 15\textdegree, 45\textdegree\space and 75\textdegree\space is utilized in the Pytorch Wavelet library \cite{pytorch_wavelets}.

The dual-tree approach ensures that the wavelet transform captures fine details from both magnitude and phase information, giving it the ability to handle directional features more effectively than the standard DWT. \\

%---------------------------------------------------------------
\subsection{Models}

% \begin{figure*}
%   \centering
%   % \hfill
%   \begin{subfigure}{0.5\textwidth}
%     \centering
%     \includegraphics[width=\textwidth]{figures/model.pdf}
%     \caption{DWT/DTCWT Model}
%     \label{model}
%   \end{subfigure}
%   % \hfill
%   \begin{subfigure}{\textwidth}
%     \centering
%     \includegraphics[width=\textwidth]{figures/model_cascaded.pdf}
%     \caption{Cascaded Model}
%     \label{model_cascaded}
%   \end{subfigure}

%   \caption{Diagrams for all proposed models.}
%   \label{fig:model_diagram}
% %    \vspace{-3mm}
% \end{figure*}

% \begin{figure*}
%     \centering
%     \includegraphics[width=1\linewidth]{figures/model_total.pdf}
%         \vspace{+0.3mm}
%     \caption{Diagram for DWT/DTCWT Model and Cascaded Model}
%     \label{fig:model_diagram}
%         \vspace{-3mm}
% \end{figure*}

To effectively reduce noise in TIR images while preserving critical structural and textural details, we propose two different denoising framework that leverages the strengths of a pretrained Stable Diffusion model. Our system reformulates the conditioning process to accommodate latent embeddings from noisy TIR inputs, and integrates a dual-domain loss function to balance both latent-level reconstruction and wavelet-domain detail preservation. The 2 proposed models are as follows:

\begin{itemize}
    \item \textbf{DWT/DTCWT Model}: Implementing DWT/DTCWT with Biorthogonal Wavelet Filter. Loss composed of Latent Loss + 2D DWT/DTCWT Loss.
    \item \textbf{Cascaded Model}: Concatenating the DWT/DTCWT Model with a Pixel-wise Loss Model for finer output
\end{itemize}

Please refer to \figref{fig:model_diagram} for detailed architecture of the models.

%-----------------------------------------------------------
\subsection{Loss}

\subsubsection{Latent Loss}
We employ a multi-component latent-space loss to ensure that the network accurately reconstructs the underlying semantic structure of the TIR data. Let $\tilde{\mathbf{x}}$ denote the input noisy TIR image, $\hat{\mathbf{x}}$ denote the predicted clean TIR image, and $\mathbf{x}$ denote the ground-truth clean TIR image in $R^{H\times W\times 3}$. First, a mean squared error (MSE) term measures the discrepancy between the predicted latent, $\hat{\mathbf{z}}$, and the corresponding ground truth latent, $\mathbf{z}$, which are both outputs of the pretrained Variational Autoencoder (VAE) model with inputs of $\hat{\mathbf{x}}$ and $\mathbf{x}$, respectively. The latent loss consists of MSE and SSIM losses, defined as

\begin{equation}
        \mathcal{L}_{\text{MSE}} = \left\| \hat{\mathbf{z}} -    \mathbf{z} \right\|^2
\end{equation}

\begin{equation}
    \mathcal{L}_{\text{SSIM}} = 1 - \text{SSIM} \left( \hat{\mathbf{z}}, \mathbf{z} \right)
\end{equation}

By balancing MSE and SSIM in the latent domain, we attain both robust noise removal and high fidelity to the underlying thermal scene. \\

%-----------------------------------------------------------

\subsubsection{2D DWT/DTCWT Loss}

%\begin{figure*}[h]
% \centering
% \includegraphics[width=\columnwidth]{figures/wavelet_comparison.pdf}
% \caption{Comparison of Various Wavelets}
% \label{wavelet_comparison}
%\end{figure*} 

In order to compute loss in the wavelet domain, we decompose both the predicted reconstruction, $\hat{\mathbf{x}}$, and the ground truth, $\mathbf{x}$, via a two-level 2D DWT/DTCWT. From the decomposition, horizontal, vertical and diagonal sub-bands $s\in\{h, v, d\}$ are utilized. We compute the mean squared error between the respective wavelet coefficients of $\hat{\mathbf{x}}$ and $\mathbf{x}$, thus directly penalizing inconsistencies in these high-frequency components:

\begin{equation}
    \mathcal{L}_{\text{wavelet}} = \sum_{s \in \{h, v, d\}} \left\| \hat{\mathbf{Y}}^s_j - \mathbf{Y}^s_j \right\|^2
\end{equation}

where $\hat{\mathbf{Y}}^s_{j}$ and $\mathbf{Y}^s_{j}$ denote the sub-band wavelet coefficients of $\hat{\mathbf{x}}$ and $\mathbf{x}$, respectively. By assigning an appropriate weight to this wavelet MSE term, we emphasize the preservation of fine details without neglecting global consistency.

Combining these objectives yields a unified loss function:

\begin{align}
\label{total_loss}
    \mathcal{L}_{\text{total}} &= 
    \underbrace{\left\| \hat{\mathbf{z}} - \mathbf{z} \right\|^2 
    + \alpha \left[ 1 - \text{SSIM}(\hat{\mathbf{z}}, \mathbf{z}) \right]}_{\text{Latent Loss}} \notag \\
    &\quad + \underbrace{\beta \sum_{s \in \{h, v, d\}} \left\| \hat{\mathbf{Y}}^s_j - \mathbf{Y}^s_j \right\|^2}_{\text{Wavelet Loss}}
\end{align}

where $\alpha$ and $\beta$ are hyperparameters balancing the contributions of the latent-space and wavelet-domain terms. On one hand, latent-space MSE and SSIM anchor the model’s outputs to the underlying scene geometry and texture distributions captured by the pretrained VAE. On the other hand, wavelet-domain constraints ensure that subtle edges and small-scale structures remain visible despite heavy noise. This hybrid objective endows our denoising framework with a robust capability to handle TIR’s wide-ranging noise patterns while maintaining crucial information for downstream robotic perception tasks. \\

\subsubsection{Pixel-wise Loss} To further refine the output of the latent model, a pixel-wise MSE and learned perceptual similarity (LPIPS) loss is implemented. Pixel-level MSE loss aligns the predicted image with the ground truth image at a per-pixel intensity level, enforcing exact pixel-value fidelity, strengthening local consistency as well as correcting minor deviations in brightness and temperature gradients. Additionally, LPIPS aligns more closely to human vision perception, focusing on high-level attributes such as texture, structure, and contrast. 
Let $\phi(\cdot)$ denotes feature extraction from the intermediate layers of a fixed image-classification backbone (e.g., AlexNet \cite{alexnet}). Formally, we define the pixel-wise loss as

\begin{equation}
        \mathcal{L}_{\text{pixel-wise}} = \left\| \hat{\mathbf{x}} -    \mathbf{x} \right\|^2 + \left\| {\mathbf{\phi(\hat{\mathbf{x}})}} -    \mathbf{\phi(\mathbf{x})} \right\|^2
\end{equation}

By augmenting the overall objective with these pixel-level and perceptual components, our final loss captures global semantics (latent space), frequency-specific details (wavelet), exact intensity alignment (pixel-wise MSE), and human- or application-relevant perceptual quality (LPIPS). This multifaceted approach mitigates oversmoothing, preserves local texture, and aligns color or brightness levels in a way that is both quantitatively accurate and perceptually coherent, ultimately enhancing TIR denoising for real-world robotic and computer vision applications.

%-----------------------------------------------------------
\subsection{Conditioning Projection Module}

Unlike text-to-image diffusion methods that use textual embeddings, our approach conditions directly on the noisy TIR latent. We encode the noisy TIR image via a pretrained VAE into a latent tensor $\mathbf{z}_{\text{noisy}} \in \mathbb{R}^{B \times 4 \times H \times W}$. The Conditioning Projection module applies adaptive average pooling and a fully connected layer to convert this latent into cross-attention embeddings $\mathbf{E}_{\text{cond}} \in \mathbb{R}^{B \times N \times D}$, which the Stable Diffusion U-Net leverages during denoising. This adaptation enables the model to capture and address the distinct noise characteristics of thermal imagery.
\renewcommand{\arraystretch}{1.2}

\begin{table*}[h]
\caption{Extensive Comparison of Our Proposed Models. Best results are highlighted in \textbf{bold} and second best are \underline{underlined}.}
\label{tab:result_compare_new}
\centering
\resizebox{0.95\textwidth}{!}{%
\begin{tabular}{c|cccc|cccc}
\hline \hline
\multirow{2}{*}{Model} & \multicolumn{4}{c|}{Loss} & \multicolumn{4}{c}{Evaluation Metric} \\
 & \phantom{0}Pixel\phantom{0} & Latent & \phantom{0}DWT\phantom{0} & DTCWT & PSNR (dB) ↑ & \phantom{0}SSIM ↑ & \phantom{0}LPIPS ↓ & \phantom{0}FID ↓ \\ \hline
Pixel-wise Loss & \checkmark &  &  &  & 23.65 & 0.7800 & 0.1769 & 69.8716 \\
Latent-Only &  & \checkmark &  &  & 26.28 & 0.8472 & 0.1767 & 46.4288 \\
Bior. DWT &  & \checkmark & \checkmark &  & 26.49 & \textbf{0.8594} & 0.1748 & 45.7763 \\
DTCWT &  & \checkmark &  & \checkmark & \textbf{27.97} & 0.8382 & \underline{0.1560} & 40.4746 \\
Bior. Cascaded & \checkmark & \checkmark & \checkmark &  & 26.32 & 0.8507 & 0.1602 & \underline{38.7875} \\
DTCWT Cascaded & \checkmark & \checkmark &  & \checkmark & \underline{26.53} & \underline{0.8556} & \textbf{0.1529} & \textbf{38.1301} \\ \hline \hline

\end{tabular}%
}
\end{table*}

\begin{table}[t]
\caption{Comparison of Denoising Models from \cite{liu2025twostage}, and Our Proposed. Best results are highlighted in \textbf{bold} and second best are \underline{underlined}.}
\label{tab:result_compare}
\centering
\resizebox{0.8\columnwidth}{!}{%
\begin{tabular}{ccc}
\hline \hline
Method                 & PSNR (dB) ↑    & SSIM ↑          \\ \hline
CLAHE \cite{setiawan2013clahe}                   & 17.23          & 0.3205          \\
NB2NB \cite{huang2021neighbor2neighbor_nb2nb} & 17.59         & 0.5552          \\
Cycle-Dehaze \cite{engin2018cycledehaze} & 13.77         & 0.7358          \\
FFDNet \cite{zhang2018ffdnet} & 17.61         & 0.5162          \\
DehazeFormer \cite{song2023dehazeformer} & 18.18 & 0.5134 \\
U$^2$D$^2$Net \cite{ding2023u2d2net} & 23.63          & \underline{0.7358}          \\
Liu et al. \cite{liu2025twostage} & \underline{24.53}          & 0.5335          \\ 
\textit{Proposed}           & \textbf{27.97} & \textbf{0.8594}\\ \hline \hline

\end{tabular}%

}
\vspace{-5mm}
\end{table}

% \begin{table}[t]
% \caption{Comparison of Denoising Models from \cite{liu2025twostage} and our proposed. Best results are highlighted in \textbf{bold} and second best are \underline{underlined}.}
% \label{tab:result_compare}
% \centering
% \resizebox{0.75\columnwidth}{!}{%
% \begin{tabular}{ccc}
% \hline \hline
% Methods                 & PSNR (dB) ↑    & SSIM ↑          \\ \hline
% CLAHE \cite{setiawan2013clahe}                   & 17.23          & 0.3205          \\
% NB2NB \cite{huang2021neighbor2neighbor_nb2nb} & 17.59         & 0.5552          \\
% Cycle-Dehaze \cite{engin2018cycledehaze} & 13.77         & 0.7358          \\
% FFDNet \cite{zhang2018ffdnet} & 17.61         & 0.5162          \\
% DehazeFormer \cite{song2023dehazeformer} & 18.18 & 0.5134 \\
% U$^2$D$^2$Net \cite{ding2023u2d2net} & 23.63          & 0.7358          \\
% Liu et al. \cite{liu2025twostage} & 24.53          & 0.5335          \\ \hline
% \textit{Pixel-wise}     & 23.65          & 0.7800          \\
% \textit{Latent Only}    & 26.28          & 0.8472          \\
% \textit{Bior. DWT}      & 26.49          & \textbf{0.8594} \\
% \textit{DTCWT}          & \textbf{27.97} & 0.8382          \\
% \textit{Cascaded Bior.} & 26.32          & 0.8507          \\
% \textit{Cascaded DTCWT} & \underline{26.53}    & \underline{0.8556}    \\ \hline \hline

% \end{tabular}%
% }
% \end{table}

\section{experiment}
\label{sec:experiment}
\subsection{Experimental Setup}  
\subsubsection{Training}
For intensive experimentation, a total of 6 models were trained as shown in \tabref{tab:result_compare_new}, with the Latent-only model imitating the original training methodology of Stable Diffusion \cite{stablediffusion}.

% \begin{itemize}
%     \item \textbf{Pixel-wise Model}: Basic fine-tuning of U-net utilizing pixel-wise MSE \& SSIM loss of ground-truth and inferred images. 
%     \item \textbf{Latent-only Model}: DWT/DTCWT Model withouth the DWT/DTCWT loss component, only utilizing latent MSE \& SSIM loss
%     \item \textbf{Bior. DWT Model}: DWT Model implementing Biorthogonal wavelet with analysis filter vanishing moment of 3.
%     \item \textbf{DTCWT Model}: DTCWT Model implementing Biorthogonal wavelet with analysis filter vanishing moment of 3.
%     \item \textbf{Bior. Cascaded Model}: Cascaded Model implementing Biorthogonal wavelet with analysis filter vanishing moment of 3.
%     \item \textbf{DTCWT Cascaded Model}: Cascaded Model implementing DTCWT with Biorthogonal wavelet with analysis filter vanishing moment of 3.
% \end{itemize}

The Stable Diffusion v1-4 model was utilized for the base foundational model, where the U-Net was fined-tuned, with the pretrained VAE model frozen. 5570 clean-noisy paired TIR image dataset from Liu et al. \cite{liu2025twostage} was utilized for the fine-tuning, with 70:30 train:test split, consistent with the experimentation from Liu et al.

All models were trained on one NVIDIA A6000 for 50 epochs with batch size of 4 and implementing Adam optimizer. Cosine annealing learning rate scheduler was utilized with starting learning rate of $1e^{-5}$ and minimum learning rate of $1e^{-7}$. For the hyperparameters from \equref{total_loss}, $j=2$ was implemented for a balance of fine-detail and global structure, $\alpha=1.0$ and $\beta = 100$ was implemented to equalize the magnitudes of losses. 

\subsubsection{Metrics}

To analyze the performance of denoising models, four metrics were used: PSNR, SSIM, LPIPS, and FID. PSNR quantifies pixel-level reconstruction quality via mean squared error, while SSIM measures structural similarities like luminance and contrast. LPIPS evaluates perceptual similarity based on deep network activations, and FID assesses realism through distributional similarity between features from real and generated images. Better performance corresponds to higher PSNR and SSIM, and lower LPIPS and FID scores.

\subsection{Evaluation}

\subsubsection{Testing on Paired Dataset}

\begin{figure*}[ht]
  \centering
  % \hfill
  \begin{subfigure}{\textwidth}
    \centering
    \includegraphics[width=\textwidth]{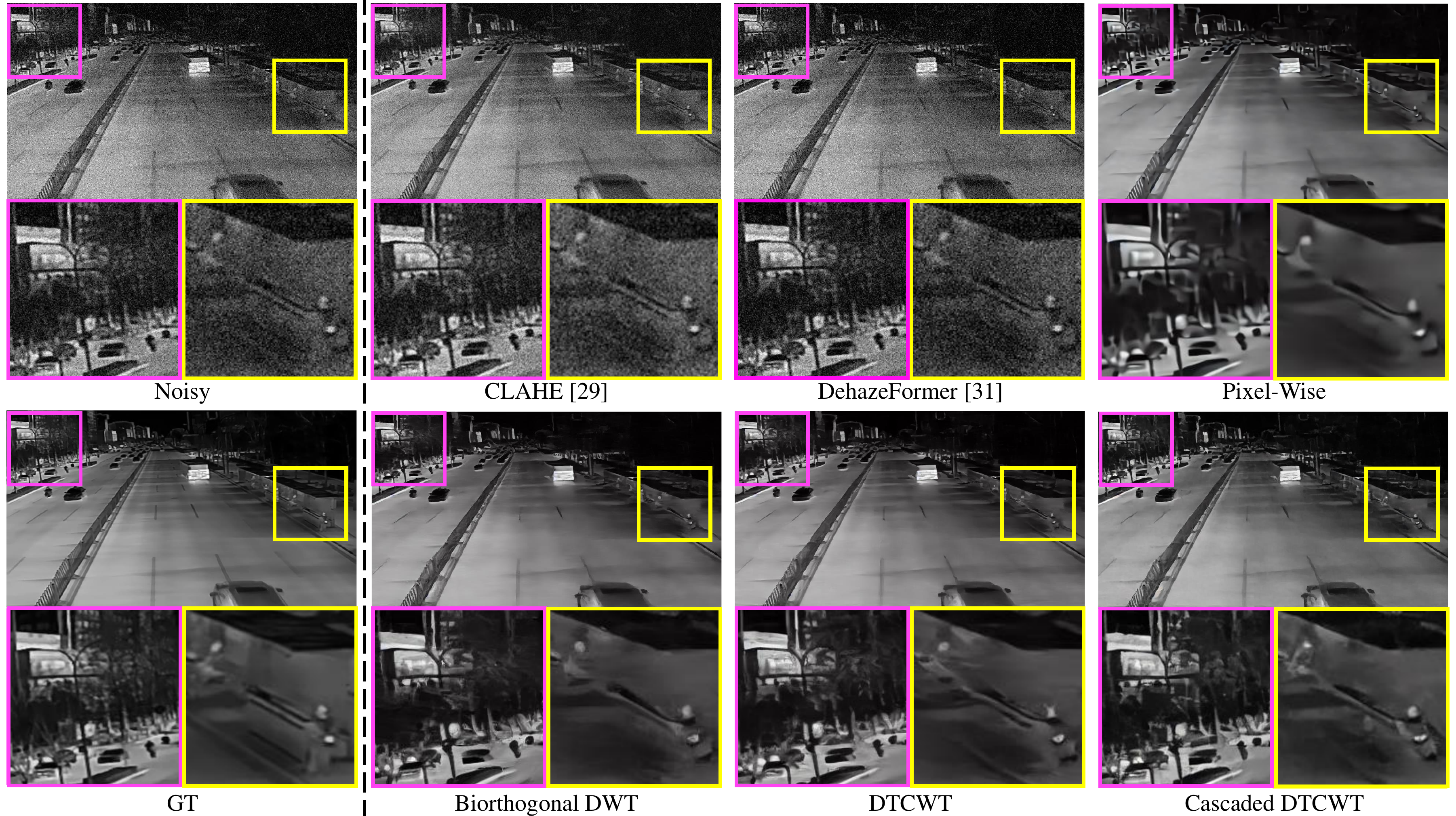}
    \caption{Outdoor Road Scene}
    \label{result1}
    \vspace{3mm}
  \end{subfigure}
  % \hfill
  \begin{subfigure}{\textwidth}
    \centering
    \includegraphics[width=\textwidth]{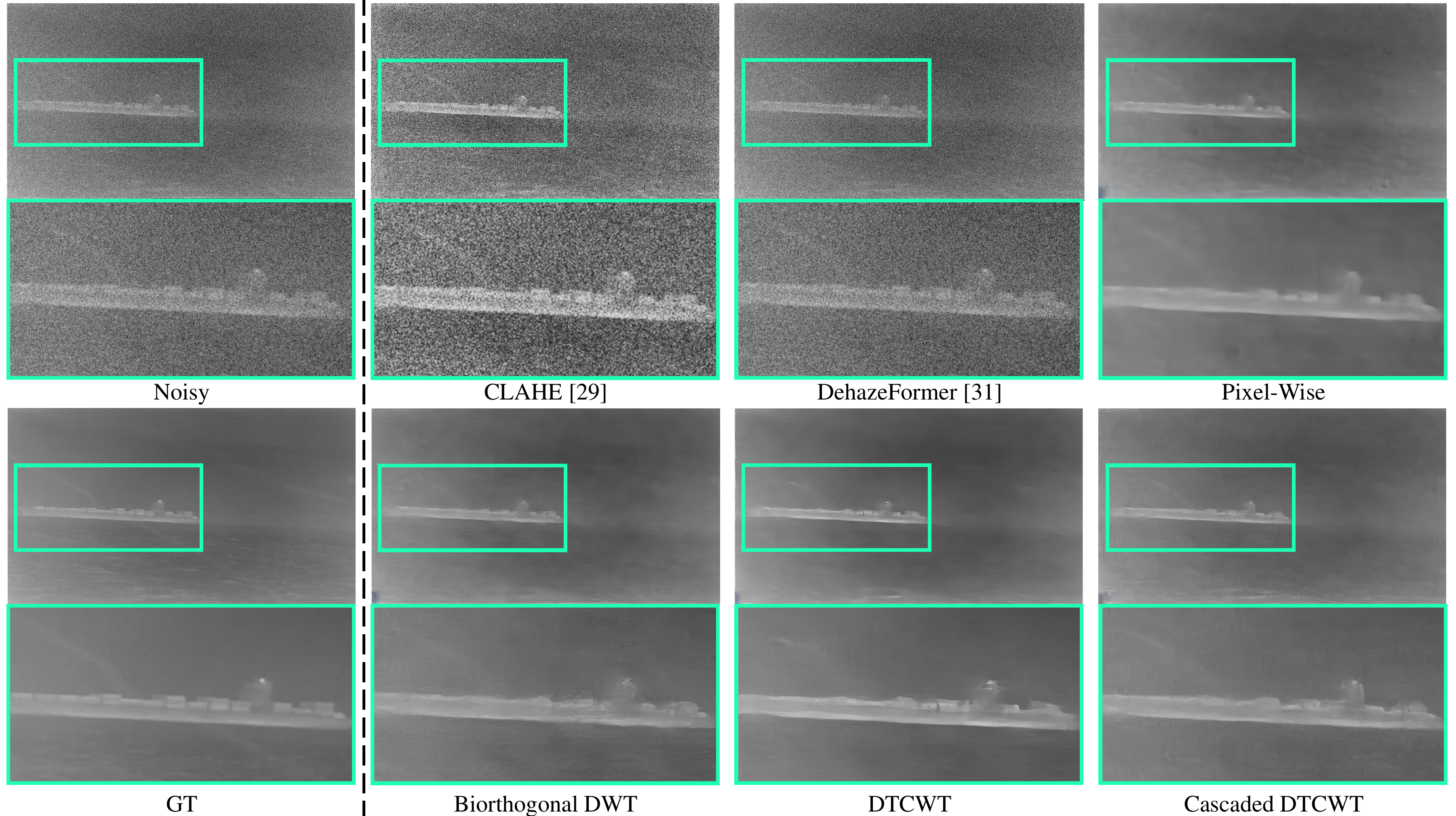}
    \caption{Maritime Ship Scene}
    \label{result2}
  \end{subfigure}
  \caption{Qualitative results on test set of two scenes comparing a traditional method, deep learning-based benchmark model, pixel-wise model and 3 best performing proposed models. Zoomed views of the colored box regions are shown in the bottom half of the images.}
  \label{fig:result12}
    \vspace{-1mm}
\end{figure*}

\begin{figure*}[t]
  \centering
  % \hfill
  \begin{subfigure}{\textwidth}
    \centering
    \includegraphics[width=\textwidth]{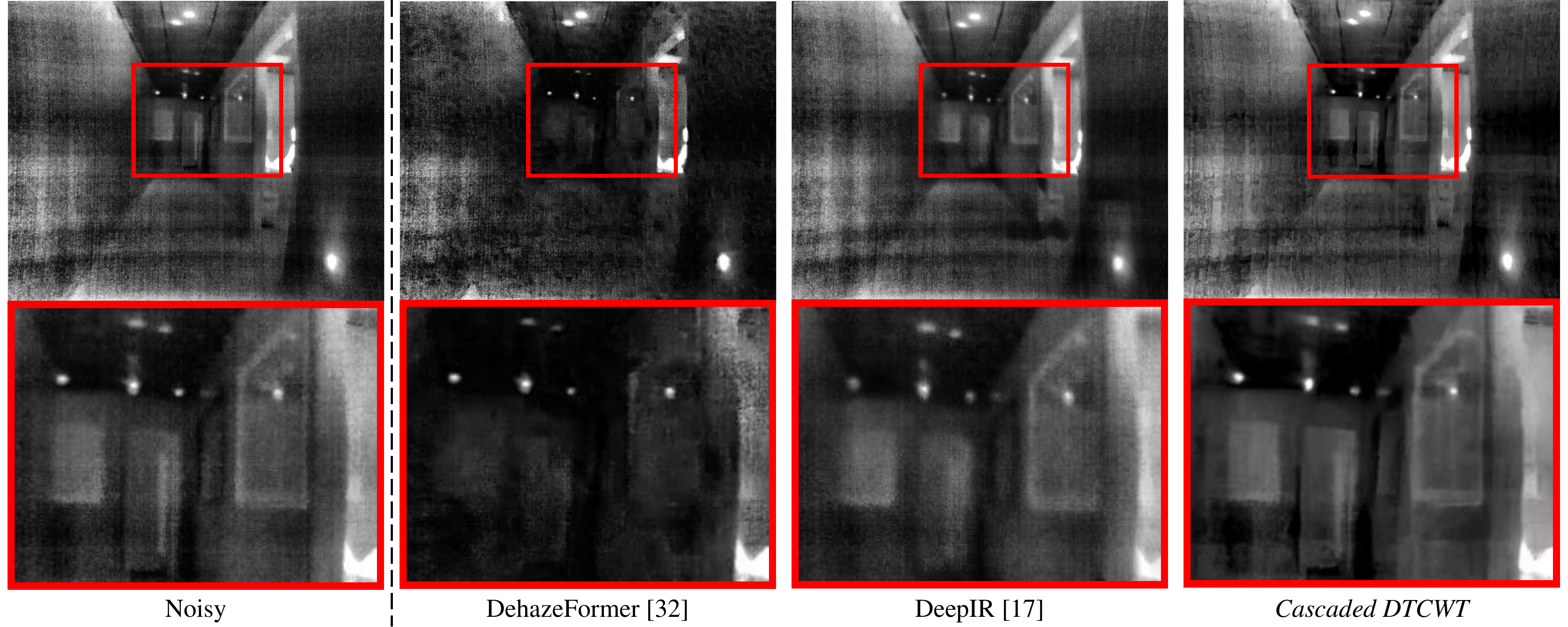}
    %\caption{Handheld Scene}
    %\vspace{+3mm}
    \label{zeroshot_result1}
  \end{subfigure}
  % \hfill
  % \begin{subfigure}{\textwidth}
  %   \centering
  %   \includegraphics[width=\textwidth]{figures/zeroshot_results2.pdf}
  %   \caption{UAV Scene}
  %   \label{zeroshot_result2}
  % \end{subfigure}
  \caption{Zero-shot experiment on OdomBeyondVision \cite{li2022odombeyondvision} dataset rescaled using Fieldscale \cite{gil2024fieldscale}. Zoomed views of the colored box regions are shown in the bottom half of the images.}
  \label{fig:zeroshot_result12}
    \vspace{-3mm}
\end{figure*}

Using the test sequence of the paired dataset, the PSNR and SSIM values were compared in \tabref{tab:result_compare} for all the models from Liu et al. \cite{liu2025twostage} and our 6 models, as LPIPS and FID values were not provided. Additionally, the 6 proposed models were extensively evaluated on all 4 proposed metrics in \tabref{tab:result_compare_new}. 

Comparing with existing results, our proposed models performed notably better in all metrics. Biorthogonal DWT model showed highest SSIM, and DTCWT model showed highest PSNR, where the cascaded DTCWT was the runner-up for both. In terms of LPIPS and FID score, Cascaded DTCWT model performed the best, with DTCWT and cascaded biorthogonal models following second, respectively. 

Compared to the Biorthogonal wavelet transform models, DTCWT models show superior performance in terms of PSNR, LPIPS and FID score, with marginal difference in SSIM. This implies that DTCWT prioritizes globally accurate pixel intensities, high perceptual fidelity, and good overall realism in distribution terms, with some lower-level or localized structural details misaligned. 

By implementing cascaded models, there is a clear improvement in LPIPS and FID score, while showing mixed results in PNSR and SSIM, implying that the model outputs more realistic and distributionally correct images at the expense of exact pixel alignment and structural similarity. This is most likely due to the LPIPS loss term in the cascaded model. 

Qualitative results in \figref{fig:result12} show clear improvements from existing and pixel-wise methods to proposed methods. Noise is clearly removed in the proposed methods, with DTCWT and cascaded DTWCT showing clearer outputs. \\

\subsubsection{Zero-shot Performance on Unseen Datasets}

Further experiments were conducted on OdomBeyondVision \cite{li2022odombeyondvision} indoor TIR image dataset rescaled via Fieldscale \cite{gil2024fieldscale}, where the images are heavily corrupted by non-uniform random noise. First images from sequences \texttt{H\_corridor\_circle\_bright\_1} and \texttt{A\_vicon2\_circle\_bright\_13} were used. Experiments were conducted on DehazeFormer \cite{song2023dehazeformer}, DeepIR \cite{saragadam2021deepir} and the proposed cascaded DTCWT model. DeepIR differs in methodology, as it requires an input of multiple images, where the physics-inspired network separates non-uniform noise using camera jitter. 20 first images of the sequence was utilized for DeepIR, as minimal movement was shown. 

Qualitative results in \figref{fig:zeroshot_result12} show much finer and clearer output for the proposed cascaded DTCWT, with existing methods failing to properly denoise the input image. Taking in consideration the extreme level of noise present in the input image, most parts which were featureless and highly corrupted were nearly untouched for most models, while the proposed method successfully denoised sections containing useful information.

\subsection{Ablation Study on Various Wavelets} \label{ssec:comparing_wavelets}

Initially, various wavelets, including the Haar \cite{haar1911theorie}, Daubechies \cite{daubechies1988orthonormal_daubechies_wavelet}, Symlet \cite{daubechies1988orthonormal_daubechies_wavelet}, Coiflet \cite{daubechies1988orthonormal_daubechies_wavelet} and Biorthogonal \cite{cohen1992biorthogonal} wavelets were tested to find the most optimal choice. In terms of implementing wavelet transforms into the loss of a denoising U-net, the most important factor would be how well it separates the noise from the input image. To evaluate this performance, the energy of wavelet coefficients is computed as below

\begin{equation}
    {E}_{s} = \left\| \mathbf{Y}^s_j \right\|^2 = \sum_{n} \sum_{m} |Y^s_j[n,m]|^2
\end{equation}

An optimal wavelet transform would induce high-frequency coefficients $(s \in \{h, v, d\})$ of large magnitude for the noisy image, while inducing very small magnitudes for the clean image, and the opposite for low-frequency coefficients $(s \in \{lo\})$. The ratios of energy of clean to noisy images were computed in \tabref{tab:wavelet_compare}, where the Biorthogonal wavelet showed best performance, followed by DTCWT. Thus, \textbf{Biorthogonal wavelet} and \textbf{DTCWT} was chosen for further experimentation.

\begin{table}[t]
\caption{Comparison of Wavelet Coefficient Energy and Noisy to Clean Energy Ratio of Various Wavelets. Best results are highlighted in \textbf{bold} and second best are \underline{underlined}.}
\label{tab:wavelet_compare}
\resizebox{\columnwidth}{!}{%
\begin{tabular}{rccccc}
\hline \hline
                              &       & $E_{lo}$ ↑              & $E_h$ ↓              & $E_v$ ↓             & $E_d$ ↓             \\ \hline
\multirow{3}{*}{Haar}         & Noisy & 2.340B    & 3.789M        & 2.585M       & 1.720M       \\
                              & Clean & 2.310B    & 2.444M        & 1.161M       & 0.239M         \\ 
                              & \textit{Ratio} & \underline{98.73\%}          & 64.51\%          & 44.94\%         & 13.93\%         \\\hline
\multirow{3}{*}{Daub.}   & Noisy & 2.596B    & 3.749M        & 2.487M       & 1.795M       \\
                              & Clean & 2.561B    & 2.230M       & 0.861M         & 0.172M         \\
                              & \textit{Ratio} & 98.63\%          & 59.48\%          & 34.65\%         & 9.62\%          \\\hline
\multirow{3}{*}{Symlet}       & Noisy & 2.630B    & 3.775M        & 2.481M       & 1.794M       \\
                              & Clean & 2.595B    & 2.248M        & 0.837M         & 0.168M         \\
                              & \textit{Ratio} & 98.68\%          & 59.55\%          & 33.74\%         & 9.40\%          \\\hline
\multirow{3}{*}{Coiflet}      & Noisy & 2.874B    & 4.137M        & 2.727M       & 2.000M       \\
                              & Clean & 2.834B    & 2.437M        & 0.922M         & 0.182M         \\
                              & \textit{Ratio} & 98.59\%    & 58.92\%          & 33.83\%         & 9.12\%    \\\hline
\multirow{3}{*}{Bior.} & Noisy & 2.464B    & 23.093M       & 21.301M      & 7.989M       \\
                              & Clean & 2.385B    & 3.443M        & 1.378M       & 0.227M         \\
                              & \textit{Ratio} & 96.81\% & \textbf{14.91\%} & \textbf{6.47\%} & \textbf{2.85\%} \\\hline
\multirow{3}{*}{DTCWT}        & Noisy & 2.341B    & 1.575M        & 1.066M         & 0.773M       \\
                              & Clean & 2.311B    & 0.916M          & 0.310M           & 0.067M          \\
                              & \textit{Ratio} & \textbf{98.74\%}          & \underline{58.14\%}    & \underline{29.09\%}    & \underline{8.77\%}         \\ \hline\hline
\end{tabular}%
\vspace{-6mm}
}

\end{table}

\section{Conclusion}

In conclusion, TIR-Diffusion implementing latent and wavelet domain losses significantly enhances TIR image quality, effectively preserving critical thermal details and exhibiting excellent generalization capabilities. These advancements contribute substantially to the practical utility and robustness of robotic perception systems in challenging real-world conditions. The proposed framework not only improves accuracy in downstream robotic tasks but also broadens the applicability of TIR imaging technology. Future work includes extending the framework greater quantities of paired datasets, allowing effective conditioning of factors including indoor and outdoor.

% \newpage

% \newpage
% \newpage

%\section*{ACKNOWLEDGMENT}
\balance
\small
\bibliographystyle{IEEEtranN} %citeauthor
\bibliography{string-short,references}

\end{document}